\documentclass{article}

     \usepackage[final]{neurips_2019}

\usepackage[utf8]{inputenc} % allow utf-8 input
\usepackage[T1]{fontenc}    % use 8-bit T1 fonts
\usepackage{hyperref}       % hyperlinks
\usepackage{url}            % simple URL typesetting
\usepackage{booktabs}       % professional-quality tables
\usepackage{amsfonts}       % blackboard math symbols
\usepackage{nicefrac}       % compact symbols for 1/2, etc.
\usepackage{microtype}      % microtypography
\usepackage{graphicx}
\usepackage{multirow}
\usepackage{booktabs}
\usepackage{caption}
\usepackage{natbib}
\setcitestyle{square,comma,numbers}

\title{Diminishing the Effect of Adversarial Perturbations via Refining Feature Representation}

\author{%
  Nader Asadi\\
  Shahid Bahonar University\\
  \texttt{naderasadi@eng.uk.ac.ir} \\
  \And
  Amir M. Sarfi \\
  Shahid Bahonar University \\
  \texttt{a.m.sarfi@gmail.com} \\
  \AND
  Mehrdad Hosseinzadeh \\
  University of Manitoba \\
  \texttt{mehrdad@cs.umanitoba.ca} \\
  \And
  Sahba Tahsini \\
  Shahid Bahonar University \\
  \texttt{tahsini.sahba@gmail.com} \\
  \And
  Mahdi Eftekhari \\
  Shahid Bahonar University \\
  \texttt{m.eftekhari@uk.ac.ir} \\
}

\begin{document}

\maketitle

\begin{abstract}
  Deep neural networks are highly vulnerable to adversarial examples, which imposes
  severe security issues for these state-of-the-art models. Many defense methods have been
  proposed to mitigate this problem. However, a lot of them depend on modification or additional training of the target model. In this work, we analytically investigate each layer's representation of non-perturbed and perturbed images and show the effect of perturbations on each of these representations. Accordingly, a method based on whitening coloring transform is proposed in order to diminish the misrepresentation of any desirable layer caused by adversaries.
  Our method can be applied to any layer of any arbitrary model without the need of any modification or additional training. Due to the fact that full whitening of the layer's representation is not easily differentiable, our proposed method is superbly robust against white-box attacks. %something about white box def%dsxs
  Furthermore, we demonstrate the strength of our method against some state-of-the-art black-box attacks.
\end{abstract}

\section{Introduction}
%add what L inf norm is where we define adversarial attacks
\label{sec:intro}
% Deep neural networks have achieved significant success in a wide variety of challenging applications.
% However, it is shown that these models are highly vulnerable to adversarial perturbations
Despite their prevalent success in a wide variety of domains, deep neural networks (DNN) become highly vulnerable when facing adversarial perturbations \cite{szegedy2013intriguing,goodfellow2014explaining}. From the security perspective, performance and robustness are of equal values; an ideal computational model ought to not only perform well, but also maintain its performance robustness against different attacks when deployed in the environment. However, 
adversarial attacks are severe threats to the robustness of DNN models. These attacks can target wide variety of domains from Google Cloud Vision \cite{ilyas2018black} to autonomous cars \cite{sitawarin2018darts}.
Therefore, it is crucial to study the security of DNN models and a new line of research is recently established to investigate and circumvent the issue \cite{barreno2006can,huang2011adversarial}.

Adversarial attacks are generated by adding minimal perturbations (which seems imperceptible to a human observer) to an image in order to mislead the target model. Due to the transferability of adversarial attacks\cite{liu2016delving}, crafted adversaries for one model can be effectively used against other models. 
Generally, adversarial attacks lay into two categories: white-box and black-box attacks. In white-box attacks, the attacker has full access to the parameters of model and uses the gradients of the model in order to generate perturbations capable of fooling the target model \cite{goodfellow2014explaining,papernot2016limitations,carlini2017towards,dong2018boosting}.
In black-box attacks, the access to the target model is limited; one have only access to the inputs and output scores of the target model \cite{narodytska2017simple}. Black-box attacks are considered a severe security issue against many practical applications since they can lead to catastrophic consequences in applications such as medical image analysis \cite{taghanaki2018vulnerability} and self-driving cars \cite{sitawarin2018darts} where the attacker might not have full access to parameters of the models. 

%Generating adversarial examples to mislead the classifier during the test phase has been investigated extensively. As a method to evaluate the effectiveness of different attacks, some constraints have been introduced to limit the additional perturbation (e.g. limiting $L_2$ or $L_\infty$ norm of the perturbations not to exceed some small value $\epsilon$). [[this paragraph doesn't make much sense to me. seems to be cut from somewhere else. can you blend it in somewhere else?]]

% Other category of attacks employ a non-gradient based approach to generate adversaries. Boundary Attack is a general-purpose hard-label attack and performs descent along the decision boundary using a rejection sampling approach \cite{brendel2017decision}.

\begin{figure*}[t!]
	\begin{center}
		\includegraphics[height=3.75cm]{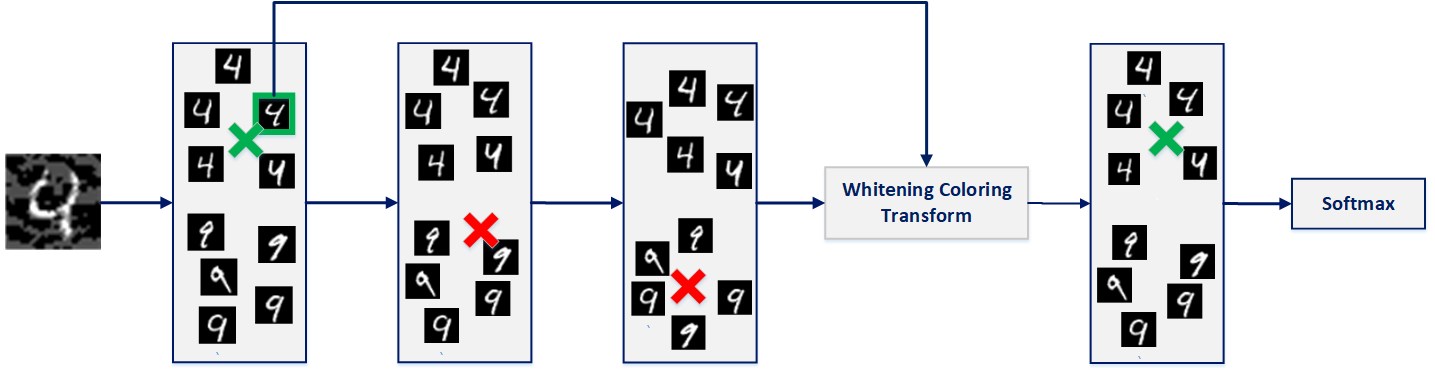}
	\end{center}
	\caption{\textbf{The intuition behind our proposed method.} We refine the feature representations of DNNs' hidden layers in order to diminish the unfavorable effects of adversarial perturbations. In order to do so, we use WCT to change the channel correlations of the adversarial example's representation in deep layers of the network.}
	\label{figure1}
\end{figure*}

Many different defense methods are designated to alleviate this unfavorable phenomenon such as pre-processing \cite{xu2017feature}, noise reduction \cite{liang2018detecting}, and using deep models \cite{samangouei2018defense,papernot2016distillation}.
These defense methods are based on the modification of deep neural networks. Even though they are effective against black-box adversarial attacks, they are still vulnerable to white-box attacks.
Likewise, another method of defense, adversarial training \cite{goodfellow2014explaining}, is shown to be effective against adversarial examples at which a number of adversaries are injected into the training data. However, In order to make our model robust using adversarial training, we need to retrain the model. Furthermore, finding appropriate adversarial examples itself is a complex challenge \cite{papernot2016limitations}. Some of the recent defense methods exploited several transformations in order to remove distortions in the image space. Guo et al. \cite{guo2017countering} investigated input transformation such as cropping, bit-depth reduction, JPEG compression, and total variance minimization while Song et al. \cite{song2018defense} performed SAAK transform. They not only deny the attacker access to easy gradients but are also effectively robust against black-box attacks. Moreover, they do not impose any additional training on the model. 
Xu et al. \cite{xu2017feature} proposed several strategies, including median smoothing and bit-depth reduction, to destruct adversarial perturbations spatially. 
In spite of all of that, many of the aforementioned methods are proven to fail against adversarial attacks\cite{carlini2017adversarial,athalye2018obfuscated}.

However,  \cite{madry2017towards} withstood many adversarial attacks, and is considered to be the current state-of-the-art defense strategy. MALADE \cite{srinivasan2018robustifying}, drives off-manifold adversarial samples towards high-density regions of the data generating the distribution of the target class by the Metroplis-adjusted Langevin algorithm (MALA) with perceptual boundary taken into account.

Xie et al. \cite{xie2019feature} pointed out that the presence of noise in feature representation of neural nets. in order to demonstrate this, they visualized the feature maps they give rise to. Inspired by this, they developed new architectures that decrease the noise in representations of adversaries and trained the resulting model end-to-end. Our work is substantially different from theirs. Firstly, we use a more analytical way to show the presence of adversaries in neural network’s representations. Moreover, in order to alleviate this noise, we do not modify the architecture nor the weights of the network leading to no need for additional training.
%maybe concat these 2 paragraphs

Due to the difficulty of performing security evaluations on defense methods, many of these methods break almost immediately after being introduced \cite{carlini2019evaluating}. For example, many defenses submitted to ICLR 2018 were broken before the review period even finished \cite{athalye2018obfuscated}. Therefore, Carlini et al. \cite{carlini2019evaluating} suggest several methods for evaluating defense strategies. We attempt to perform each item in their evaluation checklist to ensure the validity of our approach.
%[[the last sentence can go to the experimental section?]]

In this work, we investigate the transformation of layers' feature representations, refining them in order to remove the effects of perturbations from the latent representations, as a method of defense against both white and black-box attacks. To the best of our knowledge, this setting has not been investigated before from this perspective. 
Specifically, we propose the use of whitening coloring transform, as a method of representation refinement, in order to defend against adversarial examples.

We begin with an analysis on the effect of adversarial perturbations on each layer's learned feature representation in DNNs where we use nearest-neighbor algorithm to evaluate the deviation of the perturbed sample from its correct class, and we show that as the input representation goes deeper in the network, it strays further from its correct class. This motivated us to come up with a method to refine feature representation in order to alleviate the effect of adversarial attacks.
%maybe concat these 2 paragraphs

\begin{figure*}[t!]
	\begin{center}
		\includegraphics[height=4cm]{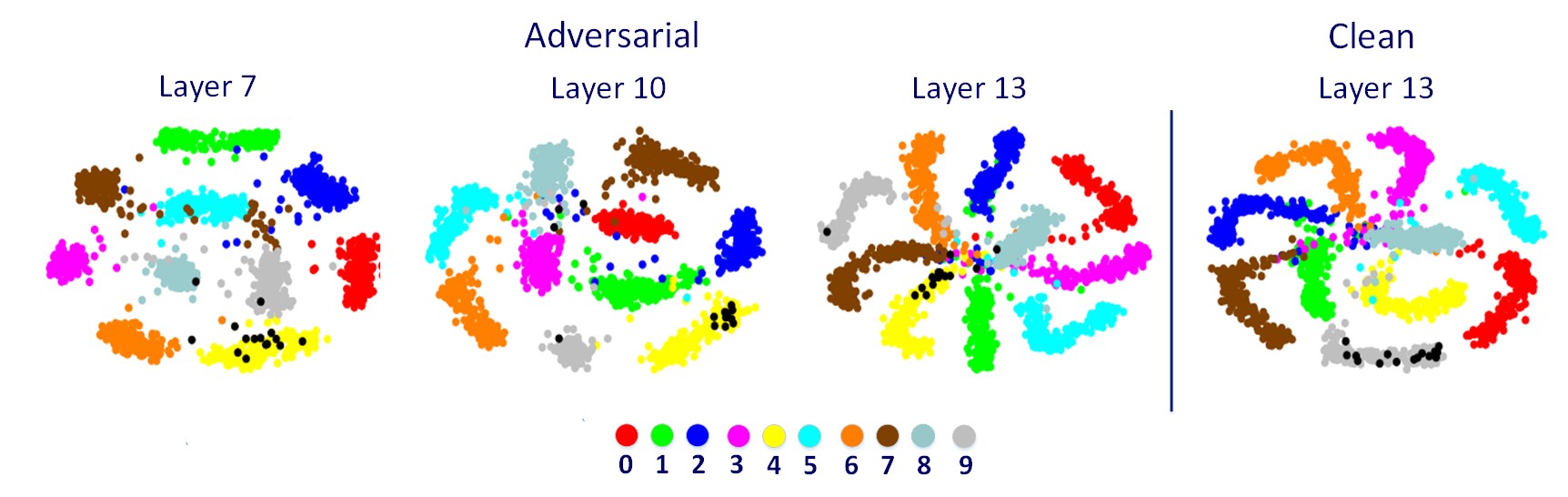}
	\end{center}
	\caption{t-SNE visualization of features learned by different layers of VGG16 network to illustrate the gradual detrimental effect of adversarial perturbations. Black dots represent perturbed samples with correct class of gray. Note that, the perturbed samples are within their correct class in the image space whereas they have deviated towards incorrect classes in deep layers.}%pls note that as it gets deeper, samples get further from their correct class
	\label{figure2}
\end{figure*}

\begin{table}[t!]
	\begin{center}
		\def\arraystretch{1}
		\begin{tabular}{l|c|c|c|c}
			&Layer4&Layer7&Layer10&Layer13\\
			\hline
			Clean&95.53&98.91&98.20&96.75\\
			Adversarial&69.19&44.51&26.35&23.29\\
		\end{tabular}
	\end{center}
	\caption{Accuracies of nearest-neighbor algorithm on feature representation of different layers on perturbed and clean samples. At deeper layers of the network, the perturbed samples' accuracy drop progressively which shows that the representations get closer to wrong classes.}
	\label{table1}
\end{table}

Our method is model-agnostic and can be applied to any application. Since full whitening of any layer's input is not easily differentiable\cite{ioffe2015batch}, our proposed method is robust against white-box attacks. As well as this, the effectiveness of our method against black-box attacks is demonstrated by reporting the results of our method against some state of the art black-box and white-box attacks.

To summarize, our work tackles the problem of adversarial perturbation in DNNs and presents an effective approach to mitigate the aforementioned problem. The contribution of this paper is thus manifold:
\begin{itemize}
	\item We analyze the impact of perturbations on feature representation of different layers in DNNs. We illustrate that as the input flows through the network, there will be a gradual shift of feature representation towards the incorrect classes.
	\item Inspired by our analysis, we propose a novel defense method to refine feature representation of DNN's hidden layers. We show that one can refine the adversarial examples' representation by whitening coloring transform and consequently, diminish the effect of adversarial perturbations drastically.
	\item Through extensive experiments, we prove the validity of our analysis and proposed method using some state-of-the-art black-box and white-box adversarial attacks.
\end{itemize}

%%%%%%%%%%%%%%%%%%%%%%%%%%%%%%%%%%%%%%%%%%%%%%%%%%%%%%%%%%%%%%
%%%%%%%%%%%%%%%%%%%% NN Analysis %%%%%%%%%%%%%%%%%%%%%%%%%
%%%%%%%%%%%%%%%%%%%%%%%%%%%%%%%%%%%%%%%%%%%%%%%%%%%%%%%%%%%%%%
\section{Analysis of the Learned Representations}\label{sec2}
In this section, we analyze the impact of adversarial attacks on the feature representation of shallow and deep layers of neural networks. 
This analysis is the foundation on which we present our proposed method in Sec. \ref{sec3}.
%We use our analysis as the basis of the next section in which we represent our method \textcolor{red}{in order to mitigate the unfavorable effect of adversarial perturbations} .

Deep neural networks learn a hierarchical set of representations and as the input flows through, the learned feature representation grows progressively abstract and each class’s representation gets further from the others, for the sake of classification. In the presence of an adversary, as the image goes deeper into the network, its corresponding feature representation seems to deviate from its correct class, constantly. At the classification layer, the deviation from the correct class seems to be enlargened enormously, throwing the model into confusion and resulting in misclassification; Hence, we focus our analysis on scrutinizing the internal representations of clean and perturbed data.

In Fig. \ref{figure2}, we use t-SNE algorithm \cite{maaten2008visualizing} to visualize the distance between the corresponding feature representations of each class. The adversarial examples are within the incorrect classes in deep layers of the network. To analyze this more accurately, we use the nearest-neighbor algorithm on those layers and calculate the accuracy in Table \ref{table1}. As the adversarial inputs get deeper into the network, their corresponding nearest-neighbor algorithm's accuracy drops indicating that these samples are getting further away from their correct class's samples. As a consequence, adversarial examples would have representations closer to incorrect classes at the classification layer, since the misrepresentation accumulates at this layer, resulting in being misclassified by the network.

Following, we will present our proposed method as a tool to alleviate any target layer's misrepresentation. In Sec. \ref{sec4}, we demonstrate the significant impact of our method on the aforementioned misrepresentation by calculating nearest-neighbor accuracy of the refined feature representations.

%%%%%%%%%%%%%%%%%%%%%%%%%%%%%%%%%%%%%%%%%%%%%%%%%%%%%%%%%%%%%%
%%%%%%%%%%%%%%%%%%%% Proposed Method %%%%%%%%%%%%%%%%%%%%%%%%%
%%%%%%%%%%%%%%%%%%%%%%%%%%%%%%%%%%%%%%%%%%%%%%%%%%%%%%%%%%%%%%

\section{Proposed Method}\label{sec3}
As discussed before, deep neural networks are extremely vulnerable to adversarial perturbation and it is because their layers misrepresent the adversaries. Also, we demonstrated that, in case of perturbations, this misrepresentation accumulates layer by layer, meaning that model's representation of the input image strays further from the space of image's correct class (in deeper layers of DNNs). Studying adversarial perturbations from this perspective leads to the idea that refining the inputs' latent representations can possibly become beneficial in order to defend effectively against such adversaries.

In what follows, we will explain the Whitening Coloring Transform (WCT) in depth and define several useful notations (section \ref{sec3.1}).  Thereupon, we will expound our novel method of refining the layers' representation in detail (section \ref{sec3.2}).

\subsection{Whitening and Coloring Transform (WCT) }\label{sec3.1}
Consider an arbitrary multivariate Gaussian random vector $X$ with arbitrary mean and covariance matrix, and $Y$, a multivariate Gaussian random vector with desirable mean and covariance matrix. WCT is the process in which the mean and covariance of $Y$ are imposed onto $X$ using two transformations details of which are as follows.

Whitening is the transformation of a target random vector $X$, to another random vector $W$ with unit diagonal covariance matrix meaning that the components of our new random vector are uncorrelated and have variances equal to 1 (this transformation is referred to as "\textit{whitening}" since the output resembles a white noise). Coloring is somehow the inverse of whitening where the desirable mean and covariance matrix are imposed to $W$.

Before whitening and coloring, both $X$ and $Y$ are centered by subtracting their mean vectors $m_x$ and $m_y$ respectively. Now consider the eigen-decompositions of covariance matrices of $X$ and $Y$ 
\begin{equation}
\Sigma_x = \Phi_x\Lambda_x\Phi^{-1}_x     \hspace{1cm}     \Sigma_y = \Phi_y\Lambda_y\Phi^{-1}_y
\end{equation}\label{eq1}
where $\Lambda$ is a diagonal matrix with eigenvalues of $\Sigma$ and $\Phi$ is the matrix of corresponding orthonormal eigenvectors such that $\Phi^{-1}=\Phi^T$.

Now $X$ is transformed into $W$ which is the uncorrelated version of $X$
\begin{equation}
W = \Phi_x\Lambda_x^{-\frac{1}{2}}\Phi^{T}_xX
\end{equation}\label{eq2}
where $\Lambda_x^{-\frac{1}{2}}\Phi^{T}_x$ decorrelates $X$ to have a covariance matrix equal to the identity matrix. $\Phi_x$ maps $W$ to the same space as $X$ since the coordinate system has been changed after transforming $X$ with $\Phi^{T}_x$, and has no effect on the covariance matrix.

Then, the coloring transform is performed on the acquired $W$:
\begin{equation}
X^\prime =  \Phi_y\Lambda_y^{\frac{1}{2}}\Phi^T_yW + m_y
\end{equation}\label{eq3}
Here, $\Phi_y\Lambda_y^{\frac{1}{2}}$ imposes the desired covariance matrix. Again, $\Phi^T_y$ just assures that $X^\prime$ ends up in the same space as $W$ since the coordinate system will be transformed by $\Phi_y$ later on and it has no effect on the covariance of $X^\prime$ as well. Finally, the desired mean vector $m_y$ will be added to the transformed representation. So now $X^\prime$, a multivariate Gaussian random vector, is obtained which has contents of $X$, with our desired mean and covariance matrix.

%\subsection{Notations}
Here, we define several useful notations. Let $I_p$ and $I_c$ denote an adversarial example and its corresponding clean image, respectively. $L(I)$ denotes correct class label of an image $I$ and let $\phi_k$ be the mapping from the image to its internal DNN representation at layer $k$ where $\phi_0$ means the image itself. Also, consider $\Gamma(\phi_k(I))$ as the image corresponding to nearest-neighbor of $\phi_k(I)$.

\begin{figure*}[t!]
	\begin{center}
		\includegraphics[height=3.5cm]{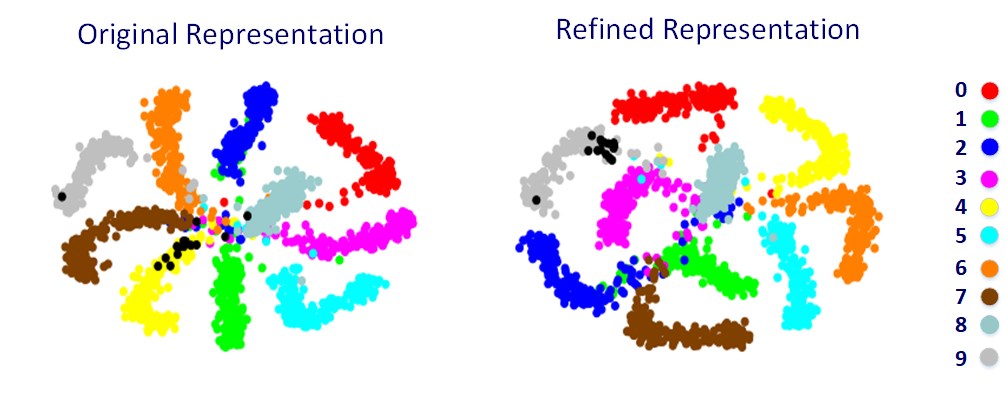}
	\end{center}
	\caption{t-SNE visualization of $13^{th}$ layer's representation. Black dots represent perturbed samples with correct class of gray. Note that, after our refinement, the representations are closer to their correct class's samples.}
	\label{figure3}
\end{figure*}

\subsection{Feature representation refinement}\label{sec3.2}
As we have debated, adversarial perturbation has an immense negative effect on layers' feature representations. This makes the neural network misunderstand the image leading to incorrect decision-making. Here we propose a method to refine the layers' representations, as a method of defense against adversarial examples. We perform WCT on the target layer's representation of the image ($\phi_k(I)$), in order to refine its features to one that is semantically in close vicinity of its true class ($L(I)$). To be precise, in order to refine an adversarial example's representation, we do the following.

For a desirable layer $k$ and an adversary $I_p$,
first, layer $k$'s feature representation of $I_p$ is whitened using Eq.\ref{eq2} such that the channels of transformed $\phi_k(I_p)$ are uncorrelated to each other and have variances of 1:
\begin{equation}
W_p = Whiten(\phi_k(I_p))
\end{equation}\label{eq8}
where $W_p$ has the same dimension and is in the same space as $I_p$ but its channels no longer have any correlation to each other. 
Table \ref{table4} represents several whitened adversarial examples. As we can see, the contents, pixels representing the most important features, are preserved while the noise is dropped tremendously. 

At this point, we may perform coloring and we would like to enforce a desirable correlation to the channels of $W_p$ for which the layer $k$'s representation of an image $I_y$ is needed. An optimal choice for $I_y$ is to choose its corresponding clean image ($I_c$) which is obviously impractical. A second optimal choice can be a random clean image whose label is the same as that of the adversary ($L(I_p)$) which as well is clearly nonviable (Table \ref{table4} shows how different images for $I_y$ affect the network).
We found that a good 
%and simple 
candidate for $I_y$ can be %as simple as 
the nearest-neighbor of $I_p$ in the training set (even better than a random image with correct label which shows that our model is not dependent on nearest-neighbor's accuracy (Table \ref{table1}) ) meaning that:
\begin{equation}
I_{y} =\Gamma(\phi_j(I_p)) \hspace{0.4cm}for\hspace{0.1cm} j \le k
\end{equation}\label{eq9}
where we choose $j=0$, the nearest-neighbor in the image space, for most of our experiments since it is more beneficial in terms of both computation and accuracy. This is due to the fact that adversaries have stronger effects on deeper layers, as stated in section \ref{sec2}). 
Now, after the calculation of $I_y$, we impose $\phi_k(I_y)$'s channels' correlations to the channels of $W$ such that the new features of layer $k$, $\phi_k^\prime(I_p)$, have the same content as $\phi_k(I_p)$ but with channel correlation of $\phi_k(I_y)$. In Table \ref{table4}, the reconstructions of $\phi_k^\prime(I_p)$ for some adversarial examples are illustrated. As one can spot, the global structures and contents are maintained while the noise has been annihilated perfectly.

The new features ($\phi_k^\prime(I_p)$) are then propagated through the rest of the network. Noticeably, our method can be applied to any number of layers simultaneously. In Fig. \ref{figure3} we show the effectiveness of our method using t-SNE visualization.

%\subsection{Method Novelties}\label{sec3.3}
%In Table\ref{table2} we show the effectiveness of our model where we put, first, one WCT on different layers and then multiple WCTs, each on a different layer. In Figure\ref{figure3}, using t-SnE, the effect of whitening coloring transform on a small set of perturbed images is demonstrated where we show that, without refinement, their representation is closer to a false class while after refiniement, they get closer to their correct class.% or inside their correct class?
%In Table\ref{table5} we show the effect of our refinement method on image denoising by decoding the feature after both whitening and coloring. In Figure\ref{figure4} we show the effect of different perturbation limits, on the accuracy of our model.  Model's accuracy decreases as the limit gets weaker (maximum allowed perturbation increases) for some attacks whereas we show that for some other attacks such as PGD, our models accuracy remains almost intact even for no constraint PGD attack.

%We argue that since the classification is almost done by these latest layers and these layers have richer corelations, since they have more channels, both extracting the correct correlation is easier and imposing the extracted correlation is of higher reward (because it leads to better classification).

%%%%%%%%%%%%%%%%%%%%%%%%%%%%%%%%%%%%%%%%%%%%%%%%%%%%%%%%%%%%%%%%%%%
%%%%%%%%%%%%%%%%%%%% Experimental Results %%%%%%%%%%%%%%%%%%%%%%%%%
%%%%%%%%%%%%%%%%%%%%%%%%%%%%%%%%%%%%%%%%%%%%%%%%%%%%%%%%%%%%%%%%%%%
\begin{table*}[t!]
	\begin{center}
		\def\arraystretch{1}
		\begin{tabular}{l|c|c|c|c|c|c|c}
			\hline
			Dataset&Attack& No Att& No Def & Layer7 & Layer10& Layer13& All 3 Layers\\
			
			\hline % <-- Midrule here
			\hline
			\multirow{5}{*}{MNIST}&FGSM$_{0.3}$ &99.52& 19.94 & 51.62 & 81.46 & 91.12& \textbf{91.46}\\
			
			&PGD$_{0.3}$&99.52 & 4.76 & 27.37 & 69.19 & 90.61& \textbf{90.98}\\
			%&$PGD_{0.8}$&99.52&2.17&10.59&60.15&89.97&\textbf{90.38}\\
			
			&CW$_{L_2}$&99.52 & 0.72 & 11.63 & 67.42 & 91.27 & \textbf{92.38}\\
			
			&MIM$_{0.3}$&99.52 & 4.32 & 17.05 & 67.51 & 84.04 & \textbf{85.48}\\
			
			&SaltnPepper&99.52 & 32.39 & - & - & 88.91 & \textbf{89.63}\\
			
			\hline % <-- Bottomrule here
			
			\multirow{3}{*}{F-MNIST}&FGSM$_{0.3}$ &91.55 & 13.48 & 11.62 & 69.55 & \textbf{84.78}& 84.69\\
			
			&PGD$_{0.3}$& 91.55 & 2.57 & 10.58 & 66.52 & \textbf{84.22}& 84.34\\
			
			%&$PGD_{0.8}$&91.55&1.61&10.02&30.86&40.64&\textbf{41.52}\\
			
			&BIM$_{0.05}$ & 91.55 & 5.31 & 10.04 & 71.91 & \textbf{84.49}& 84.16\\
			
			\hline % <-- Bottomrule here
			
			\multirow{3}{*}{Fonts}&FGSM$_{0.3}$ &91.35 & 5.33 & 11.53 & 80.85 & \textbf{85.86}& 85.72\\
			
			&PGD$_{0.3}$& 91.35 & 3.24 & 15.66 & 80.66 & 85.42 & \textbf{85.89}\\
			
			%&$PGD_{0.8}$&91.35&0.76&5.12&76.07&80.52&\textbf{81.08}\\
			
			&BIM$_{0.05}$ & 91.35 & 4.52 & 56.74 & 82.05 & 85.32& \textbf{85.72}\\
			
			\hline % <-- Bottomrule here
		\end{tabular}
	\end{center}
	\caption{Comparison of the accuracies of our method under different settings. Each column shows which layer we have put the WCT on.}
	\label{table2}
\end{table*}

\begin{table*}[t!]
	\begin{center}
		\def\arraystretch{1}
		\begin{tabular}{l|c|c|c|c||c|c|c|c}
			\hline
			&\multicolumn{4}{c||}{FGSM$_{0.3}$}&\multicolumn{4}{c}{PGD$_{0.3}$}\\
			\hline
			&Layer7&Layer10&Layer13&model&Layer7&Layer10&Layer13&model\\
			\hline
			\hline
			Adv.&44.51&26.35&23.29&19.94&18.45&7.61&4.82&3.89\\
			Layer7&88.50&60.99&54.54&53.57&58.68&20.63&18.89&18.11\\
			Layer10&-&90.12&76.54&81.46&-&73.13&53.78&63.82\\
			Layer13&-&-&90.94&91.12&-&-&80.24&90.61\\
			All 3 Layers&88.50&91.98&91.94&\textbf{91.46}&58.68&83.34&90.57&\textbf{90.98}\\
			\hline
		\end{tabular}
	\end{center}
	\caption{The effect of using our refinement on different layers of the network. Each row shows which layer we have put the WCT on and each column shows the nearest-neighbor's accuracy on different layers' representations. First row shows the accuracies on the vanilla network and last column shows the accuracy of the whole model.}
	\label{table3}
\end{table*}

\section{Experiments and Results}\label{sec4}

We empirically evaluate our feature representation refinement method. We test our method in different settings (\textit{i.e.} applying it on different layers) and investigate its refinement effect. Based on the evaluation checklist introduced in \cite{carlini2019evaluating}, we verify the correctness and effectiveness of our model. Our experiments are conducted on three publicly available datasets: MNIST\cite{lecun1998mnist}, Fashion-MNIST\cite{xiao2017fashion}, and Letters\cite{cohen2017emnist}.

\textbf{Implementation details.} For all experiments, we use VGG16, and in order to visualize the denoizing effect of our method, VGG19 is used. Furthermore, for all experiments comprising nearest-neighbor, we use only 200 samples for each class randomly selected from almost $16.6\%$ of the training set in order to have the minimum computational overhead. Likewise, for all of our t-SNE visualizations, we use 200 images of each class as the clean data and 15 adversarial examples.

%\subsection{Quantitative Analysis of Refinement Effects}
\subsection{Ablation analysis}
\subsubsection{WCT at different layers}
WTC technique has different effects depending on which layer it is applied upon. To investigate this, WTC is applied on different layers of the target network and the results are presented in Table \ref{table1}. In section \ref{sec2}, we discussed that in the presence of adversaries, latent features of images strays further from its correct class at deeper layers of the network. In table \ref{table3}, we demonstrate the proficiency of our method on feature representation refinement. In order to do so, we put our proposed method on different layers of the network, and by a similar experiment to that of section \ref{sec2}, we seek the refined representation's nearest neighbor accuracy in the subsequent layers. As one can see, the improvement after WCT is considerable. In fact, the deeper we do the WCT, the better results we achieve (when using only one WCT). We argue that this is due to the fact that deeper layers' representations are more sparse (for the sake of classification) and their values have stronger correlations to each other. Therefore, imposing correct correlations to a perturbed sample at these layers will have a higher reward. Also, performing the transform on all layers of the network has a considerable effect on every layer and subsequently, the accuracy of the model. We suggest using our method on last layers if multiple WCTs is not possible (since performing the transformation on deep layers has acceptable performance and much less computational overhead).

\begin{table*}[t!]
	\begin{center}
		\def\arraystretch{1}
		\begin{tabular}{l|c|c|c|c|c|c}
			\hline
			&image domain&Layer7&Layer10&Layer13&correct class&ground truth\\
			\hline
			Accuracy&91.36&37.11&21.14&19.64&90.54&94.27\\
			\hline
		\end{tabular}
	\end{center}
	\caption{Comparison on the effectiveness of different choices for WCT's clean image. Note that, throughout this experiment, WCT is performed on $13^{th}$ layer of VGG16.}
	\label{table4}
\end{table*}

%\begin{table*}[t!]
%	\begin{center}
%		\def\arraystretch{1}
%		\begin{tabular}{l|c c c c c}
%			
%			\hline
%			Attack&CNN&Madry&CNN + MALADE&Madry + MALADE&Ours\\
%			\hline
%			SaltnPepper&36.49&41.61&80.41&80.72&\textbf{88.9}\\
%			Boundary Attack&32.39&1.10&93.79&95.80&\\
%			\hline
%			
%		\end{tabular}
%	\end{center}
%	\caption{Performance comparison against non-gradient based attacks on MNIST dataset.}
%	\label{table5}
%\end{table*}

\begin{table*}[t!]
	\parbox{.45\linewidth}{
		\centering
		\tabcolsep=0.15cm
		\begin{tabular}{l|c c c c c}
			
			\hline
			Defenses&FGSM&BIM&MIM&PGD&C\&W\\
			\hline
			\hline
			&\multicolumn{5}{c}{$\epsilon$ = 0.2 , $c$ = 1}\\
			\hline
			Baseline&12.9&12.2&16.1&6.3&30.6\\
			Yu&70.3&77.1&-&-&79.1\\
			Ross&60.4&32.1&29.3&44.2&75.3\\
			Pang&52.8&73.6&77.5&41.0&78.1\\
			Mustafa&77.9&77.3&82.0&80.2&91.2\\
			Ours&\textbf{91.2}&\textbf{92.1}&\textbf{89.4}&\textbf{92.2}&\textbf{92.3}\\
			\hline
			&\multicolumn{5}{c}{$\epsilon$ = 0.3 , $c$ = 10}\\
			\hline
			Mustafa&85.2&81.9&82.8&80.8&83.5\\
			Ours&\textbf{91.3}&\textbf{89.3}&\textbf{84.0}&\textbf{90.6}&\textbf{91.2}\\
			\hline
			
		\end{tabular}
		\caption{Comparison on MNIST. For our model, we report results with transformation on the last convolutional layer.}
		\label{table6}
	}
	\hfill
	\parbox{.45\linewidth}{
		\centering
		\tabcolsep=0.15cm
		\begin{tabular}{l|c c}
			
			\hline
			Defenses&FGSM&MIM\\
			\hline
			Baseline&53.34&1.13\\
			Bit Depth Reduction (1-bit)&86.96&89.02\\
			Median Smoothing (2x2)&57.37&28.63\\
			Median Smoothing (3x3)&54.44&27.19\\
			JPEG (Q=10)&61.03&44.19\\
			JPEG (Q=75)&53.75&17.65\\
			Ours&\textbf{92.3}&\textbf{92.0}\\
			\hline
			
		\end{tabular}
		\caption{Comparison with some pre-processing methods on MNIST. The perturbation magnitude($\epsilon$) of 0.1 is used for both attacks.}
		\label{table7}
	}
\end{table*}

\subsubsection{Nearest-neighbor from different layers}
Table \ref{table4} presents a comparison of the effectiveness of choosing WCT's clean image $I_y$ (using the nearest-neighbor algorithm) from different layers of the model in order to find an appropriate mean and covariance. It is shown that the best practical choice for $I_y$ is the nearest-neighbor of the adversary in the image space. Likewise, choosing the clean image from shallower layers has much better performance since adversarial attacks have a huge impact on the representation of deeper layers. This stems from the fact that the misrepresentation caused by adversarial perturbations accumulates at later layers, as discussed in section \ref{sec2}.
	
%\begin{table}[t!]
%	\begin{center}
%		\def\arraystretch{1}
%		\begin{tabular}{l|c c c c c}
%			
%			\hline
%			\multirow{2}{*}{Defense}&\multirow{2}{*}{Clean}&FGSM&FGSM&\multirow{2}{*}{PGD}&\multirow{2}{*}{C\&W}\\
%			&&(0.05)&(0.15)&&\\
%			\hline
%			Baseline&90.9&60.7&18.6&0.0&0.0\\
%			FGSM$_{0.2}$&\multirow{2}{*}{90.3}&\multirow{2}{*}{81.1}&\multirow{2}{*}{58.5}&\multirow{2}{*}{2.0}&\multirow{2}{*}{54.4}\\
%			AdvTrain&&&&&\\
%			FGSM$_{0.4}$&\multirow{2}{*}{89.0}&\multirow{2}{*}{80.5}&\multirow{2}{*}{76.0}&\multirow{2}{*}{14.9}&\multirow{2}{*}{82.3}\\
%			AdvTrain&&&&&\\
%			PGD AdvTrain&83.5&81.5&77.7&67.3&84.9\\
%			Distillation&88.1&80.4&80.2&80.1&0.0\\
%			Defense-GAN&83.5&78.9&69.0&49.8&37.9\\
%			BCGAN&85.4&83.8&83.8&83.6&\textbf{85.9}\\
%			\hline
%			Ours&89.2&\textbf{84.9}&\textbf{84.5}&\textbf{84.1}&85.6\\
%			\hline
%		\end{tabular}
%	\end{center}
%	\caption{Comparison with different defense methods against various attacks on the Fashion-MNIST dataset.}
%	\label{table9}
%\end{table}

\subsection{Robustness of WCT refinement}
First off, in table \ref{table2}, we can see that the proposed model, against the vanilla model, has an acceptable accuracy drop of almost $1.5\%$ on clean data. Thereon, the effectiveness of our method (placed at different layers) is evaluated against different attacks (FGSM\cite{goodfellow2014explaining}, BIM\cite{feinman2017detecting}, PGD\cite{madry2017towards}, and CW\cite{carlini2017towards}  and MIM\cite{dong2018boosting}) with different distortions on the three datasets mentioned before. The attacks' subscripts denote the $L_\infty$ constraint imposed onto them except for CW whose subscript denotes its constant. These attacks were performed in a black-box manner where the adversaries are created against a pre-trained vanilla model (VGG16) and transferred to the same model with our method in different settings. This is because of the fact that it is near to impossible to pass the gradient through our method and the randomness of the nearest-neighbor (since there is no guarantee that the selected nearest-neighbor would remain the same for the clean and adversarial samples).  Lastly, in fig. \ref{figure4} we demonstrate that, as the distortion bound ($\epsilon$) increases, the robustness of our method is reduced, monotonically. Furthermore, for unlimited distortion ($L_\infty = \infty$) it shows near to $0\%$ robustness which is expected of any defense model \cite{carlini2019evaluating}. Also, the reconstruction of denoised feature embeddings of adversaries can be found in Appendix \ref{appendix}.

\begin{table*}[t!]
	\parbox{.5\linewidth}{
		\begin{center}
			\tabcolsep=0.15cm
			\begin{tabular}{l|c c c c c}
				
				\hline
				\multirow{2}{*}{Defense}&\multirow{2}{*}{Clean}&FGSM&FGSM&\multirow{2}{*}{PGD}&\multirow{2}{*}{C\&W}\\
				&&(0.05)&(0.15)&&\\
				\hline
				Baseline&90.9&60.7&18.6&0.0&0.0\\
				FGSM$_{0.2}$&\multirow{2}{*}{90.3}&\multirow{2}{*}{81.1}&\multirow{2}{*}{58.5}&\multirow{2}{*}{2.0}&\multirow{2}{*}{54.4}\\
				AdvTrain&&&&&\\
				FGSM$_{0.4}$&\multirow{2}{*}{89.0}&\multirow{2}{*}{80.5}&\multirow{2}{*}{76.0}&\multirow{2}{*}{14.9}&\multirow{2}{*}{82.3}\\
				AdvTrain&&&&&\\
				PGD AdvTrain&83.5&81.5&77.7&67.3&84.9\\
				Distillation&88.1&80.4&80.2&80.1&0.0\\
				Defense-GAN&83.5&78.9&69.0&49.8&37.9\\
				BCGAN&85.4&83.8&83.8&83.6&\textbf{85.9}\\
				\hline
				Ours&89.2&\textbf{84.9}&\textbf{84.5}&\textbf{84.1}&85.6\\
				\hline
			\end{tabular}
		\end{center}
		\caption{Comparison with different defense methods against various attacks on the Fashion-MNIST.}
		\label{table9}	
	}
	\hfill
	\parbox{.375\linewidth}{
		\begin{center}
			\includegraphics[height=3.5cm]{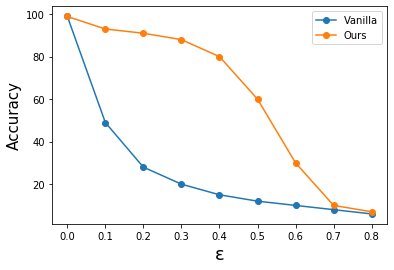}
		\end{center}
	\captionof{figure}{Robustness of our model against various perturbation bounds ($\epsilon$) on FGSM.}
		\label{figure4}	
	}
\end{table*}

\subsection{Comparison with prior defences}
We compare the robustness of our method against the majority of existing defenses. Firstly, we use gradient-based attacks done in a black-box manner. we compare the robustness of our method against several adversarial perturbation based defenses on the MNIST dataset. We used the accuracies reported by \cite{mustafa2019adversarial} and we find our proposed method to outperform the others (Table \ref{table6}). As the second category of prior defenses, we consider several transformation-based methods. These models apply various transformations on the input in order to denoise the adversary. Table \ref{table7} demonstrates the prominence of our method against these defenses footnote{In order to report defense accuracy in this table, we used the corresponding Advertorch implementations.}.
Table \ref{table9} presents our experiments on F-MNIST dataset. 
Here, we use the reported accuracies in \cite{sun2019enhancing}, and again, our method beats the state-of-the-art under most settings.

\section{Conclusion and Future Works}
In this work, we studied the effect of adversarial perturbations on feature representations of
hidden layers and we proposed a novel method to tackle this problem. Our method tries to refine the adversaries' latent representations and is independent of any modification or additional training of the model. Moreover, our method is
robust against white-box attacks since computing the gradients of whitening coloring transform is hard and has huge computational overhead. As well as this, we demonstrated the effectiveness of our model against several state-of-the-art black-box attacks.

Although the experimental results show that the proposed method is a simple yet effective model-agnitic tool against adversarial attack, it is by no means perfect; utilizing  nearest-neighbor algorithm to find a clean image from training data for refinement of deep layers not only introduces computational overhead, but it also can fall short for more complex problems and datasets, indicating the demand to invest more in this research topic. 

% For future work, we aim to solve the problem by omitting either the need of clean image completely or finding an alternative to nearest-neighbor which can be applied to complex datasets as well.

\bibliographystyle{unsrt}  
%\bibliography{refrences}

\begin{thebibliography}{10}

\bibitem{szegedy2013intriguing}
Christian Szegedy, Wojciech Zaremba, Ilya Sutskever, Joan Bruna, Dumitru Erhan,
  Ian Goodfellow, and Rob Fergus.
\newblock Intriguing properties of neural networks.
\newblock {\em arXiv preprint arXiv:1312.6199}, 2013.

\bibitem{goodfellow2014explaining}
Ian~J Goodfellow, Jonathon Shlens, and Christian Szegedy.
\newblock Explaining and harnessing adversarial examples.
\newblock {\em arXiv preprint arXiv:1412.6572}, 2014.

\bibitem{ilyas2018black}
Andrew Ilyas, Logan Engstrom, Anish Athalye, and Jessy Lin.
\newblock Black-box adversarial attacks with limited queries and information.
\newblock {\em arXiv preprint arXiv:1804.08598}, 2018.

\bibitem{sitawarin2018darts}
Chawin Sitawarin, Arjun~Nitin Bhagoji, Arsalan Mosenia, Mung Chiang, and
  Prateek Mittal.
\newblock Darts: Deceiving autonomous cars with toxic signs.
\newblock {\em arXiv preprint arXiv:1802.06430}, 2018.

\bibitem{barreno2006can}
Marco Barreno, Blaine Nelson, Russell Sears, Anthony~D Joseph, and J~Doug
  Tygar.
\newblock Can machine learning be secure?
\newblock In {\em Proceedings of the 2006 ACM Symposium on Information,
  computer and communications security}, pages 16--25. ACM, 2006.

\bibitem{huang2011adversarial}
Ling Huang, Anthony~D Joseph, Blaine Nelson, Benjamin~IP Rubinstein, and
  JD~Tygar.
\newblock Adversarial machine learning.
\newblock In {\em Proceedings of the 4th ACM workshop on Security and
  artificial intelligence}, pages 43--58. ACM, 2011.

\bibitem{liu2016delving}
Yanpei Liu, Xinyun Chen, Chang Liu, and Dawn Song.
\newblock Delving into transferable adversarial examples and black-box attacks.
\newblock {\em arXiv preprint arXiv:1611.02770}, 2016.

\bibitem{papernot2016limitations}
Nicolas Papernot, Patrick McDaniel, Somesh Jha, Matt Fredrikson, Z~Berkay
  Celik, and Ananthram Swami.
\newblock The limitations of deep learning in adversarial settings.
\newblock In {\em 2016 IEEE European Symposium on Security and Privacy
  (EuroS\&P)}, pages 372--387. IEEE, 2016.

\bibitem{carlini2017towards}
Nicholas Carlini and David Wagner.
\newblock Towards evaluating the robustness of neural networks.
\newblock In {\em 2017 IEEE Symposium on Security and Privacy (SP)}, pages
  39--57. IEEE, 2017.

\bibitem{dong2018boosting}
Yinpeng Dong, Fangzhou Liao, Tianyu Pang, Hang Su, Jun Zhu, Xiaolin Hu, and
  Jianguo Li.
\newblock Boosting adversarial attacks with momentum.
\newblock In {\em Proceedings of the IEEE conference on computer vision and
  pattern recognition}, pages 9185--9193, 2018.

\bibitem{narodytska2017simple}
Nina Narodytska and Shiva Kasiviswanathan.
\newblock Simple black-box adversarial attacks on deep neural networks.
\newblock In {\em 2017 IEEE Conference on Computer Vision and Pattern
  Recognition Workshops (CVPRW)}, pages 1310--1318. IEEE, 2017.

\bibitem{taghanaki2018vulnerability}
Saeid~Asgari Taghanaki, Arkadeep Das, and Ghassan Hamarneh.
\newblock Vulnerability analysis of chest x-ray image classification against
  adversarial attacks.
\newblock In {\em Understanding and Interpreting Machine Learning in Medical
  Image Computing Applications}, pages 87--94. Springer, 2018.

\bibitem{xu2017feature}
Weilin Xu, David Evans, and Yanjun Qi.
\newblock Feature squeezing: Detecting adversarial examples in deep neural
  networks.
\newblock {\em arXiv preprint arXiv:1704.01155}, 2017.

\bibitem{liang2018detecting}
Bin Liang, Hongcheng Li, Miaoqiang Su, Xirong Li, Wenchang Shi, and XiaoFeng
  Wang.
\newblock Detecting adversarial image examples in deep neural networks with
  adaptive noise reduction.
\newblock {\em IEEE Transactions on Dependable and Secure Computing}, 2018.

\bibitem{samangouei2018defense}
Pouya Samangouei, Maya Kabkab, and Rama Chellappa.
\newblock Defense-gan: Protecting classifiers against adversarial attacks using
  generative models.
\newblock {\em arXiv preprint arXiv:1805.06605}, 2018.

\bibitem{papernot2016distillation}
Nicolas Papernot, Patrick McDaniel, Xi~Wu, Somesh Jha, and Ananthram Swami.
\newblock Distillation as a defense to adversarial perturbations against deep
  neural networks.
\newblock In {\em 2016 IEEE Symposium on Security and Privacy (SP)}, pages
  582--597. IEEE, 2016.

\bibitem{guo2017countering}
Chuan Guo, Mayank Rana, Moustapha Cisse, and Laurens van~der Maaten.
\newblock Countering adversarial images using input transformations.
\newblock {\em arXiv preprint arXiv:1711.00117}, 2017.

\bibitem{song2018defense}
Sibo Song, Yueru Chen, Ngai-Man Cheung, and C-C~Jay Kuo.
\newblock Defense against adversarial attacks with saak transform.
\newblock {\em arXiv preprint arXiv:1808.01785}, 2018.

\bibitem{carlini2017adversarial}
Nicholas Carlini and David Wagner.
\newblock Adversarial examples are not easily detected: Bypassing ten detection
  methods.
\newblock In {\em Proceedings of the 10th ACM Workshop on Artificial
  Intelligence and Security}, pages 3--14. ACM, 2017.

\bibitem{athalye2018obfuscated}
Anish Athalye, Nicholas Carlini, and David Wagner.
\newblock Obfuscated gradients give a false sense of security: Circumventing
  defenses to adversarial examples.
\newblock {\em arXiv preprint arXiv:1802.00420}, 2018.

\bibitem{madry2017towards}
Aleksander Madry, Aleksandar Makelov, Ludwig Schmidt, Dimitris Tsipras, and
  Adrian Vladu.
\newblock Towards deep learning models resistant to adversarial attacks.
\newblock {\em arXiv preprint arXiv:1706.06083}, 2017.

\bibitem{srinivasan2018robustifying}
Vignesh Srinivasan, Arturo Marban, Klaus-Robert Müller, Wojciech Samek, and
  Shinichi Nakajima.
\newblock Robustifying models against adversarial attacks by langevin dynamics,
  2018.

\bibitem{xie2019feature}
Cihang Xie, Yuxin Wu, Laurens van~der Maaten, Alan~L Yuille, and Kaiming He.
\newblock Feature denoising for improving adversarial robustness.
\newblock In {\em Proceedings of the IEEE Conference on Computer Vision and
  Pattern Recognition}, pages 501--509, 2019.

\bibitem{carlini2019evaluating}
Nicholas Carlini, Anish Athalye, Nicolas Papernot, Wieland Brendel, Jonas
  Rauber, Dimitris Tsipras, Ian Goodfellow, and Aleksander Madry.
\newblock On evaluating adversarial robustness.
\newblock {\em arXiv preprint arXiv:1902.06705}, 2019.

\bibitem{ioffe2015batch}
Sergey Ioffe and Christian Szegedy.
\newblock Batch normalization: Accelerating deep network training by reducing
  internal covariate shift.
\newblock {\em arXiv preprint arXiv:1502.03167}, 2015.

\bibitem{maaten2008visualizing}
Laurens van~der Maaten and Geoffrey Hinton.
\newblock Visualizing data using t-sne.
\newblock {\em Journal of machine learning research}, 9(Nov):2579--2605, 2008.

\bibitem{lecun1998mnist}
Yann LeCun.
\newblock The mnist database of handwritten digits.
\newblock {\em http://yann. lecun. com/exdb/mnist/}, 1998.

\bibitem{xiao2017fashion}
Han Xiao, Kashif Rasul, and Roland Vollgraf.
\newblock Fashion-mnist: a novel image dataset for benchmarking machine
  learning algorithms.
\newblock {\em arXiv preprint arXiv:1708.07747}, 2017.

\bibitem{cohen2017emnist}
Gregory Cohen, Saeed Afshar, Jonathan Tapson, and Andr{\'e} van Schaik.
\newblock Emnist: an extension of mnist to handwritten letters.
\newblock {\em arXiv preprint arXiv:1702.05373}, 2017.

\bibitem{feinman2017detecting}
Reuben Feinman, Ryan~R Curtin, Saurabh Shintre, and Andrew~B Gardner.
\newblock Detecting adversarial samples from artifacts.
\newblock {\em arXiv preprint arXiv:1703.00410}, 2017.

\bibitem{mustafa2019adversarial}
Aamir Mustafa, Salman Khan, Munawar Hayat, Roland Goecke, Jianging Shen, and
  Ling Shao.
\newblock Adversarial defense by restricting the hidden space of deep neural
  networks.
\newblock {\em arXiv preprint arXiv:1904.00887}, 2019.

\bibitem{sun2019enhancing}
Ke~Sun, Zhanxing Zhu, and Zhouchen Lin.
\newblock Enhancing the robustness of deep neural networks by boundary
  conditional gan.
\newblock {\em arXiv preprint arXiv:1902.11029}, 2019.

\bibitem{hendrycks2018benchmarking}
Dan Hendrycks and Thomas~G Dietterich.
\newblock Benchmarking neural network robustness to common corruptions and
  surface variations.
\newblock {\em arXiv preprint arXiv:1807.01697}, 2018.

\bibitem{brendel2017decision}
Wieland Brendel, Jonas Rauber, and Matthias Bethge.
\newblock Decision-based adversarial attacks: Reliable attacks against
  black-box machine learning models.
\newblock {\em arXiv preprint arXiv:1712.04248}, 2017.

\end{thebibliography}

\newpage

\section{Appendix}
\subsection{Feature Reconstructions}\label{appendix}

\begin{figure*}[h!]
	\begin{center}
		\includegraphics[height=12.5cm]{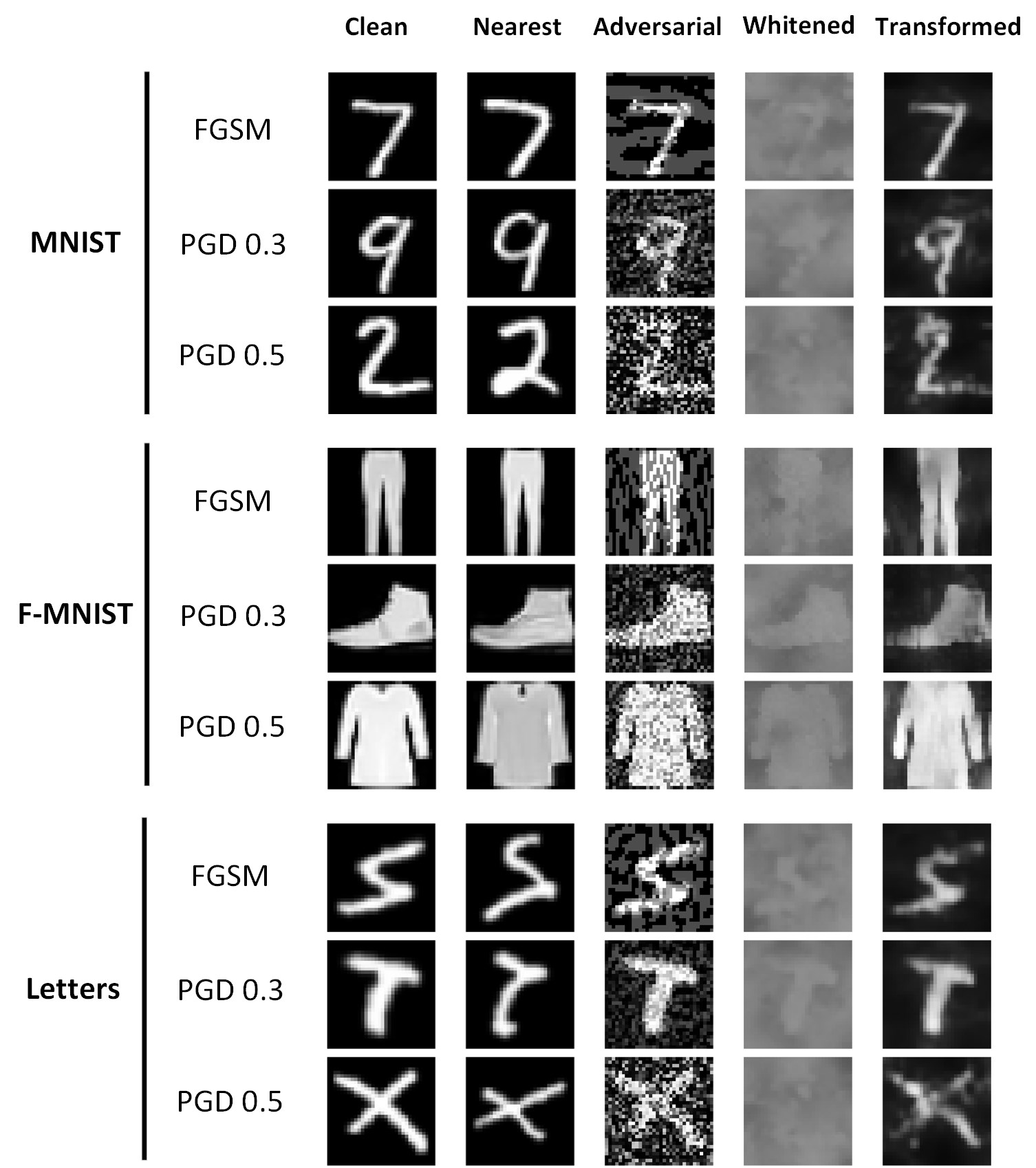}
	\end{center}
	\caption{Visualizing the refined and whitened representations of several adversarial examples.}
	\label{figure5}
\end{figure*}

In figure \ref{figure5}, the effectiveness of our method is visualized. To do so, we use VGG19 auto-encoder and use WCT on its encoder's $16^{th}$ layer. Here, the corresponding clean image, the chosen nearest-neighbor, and the adversarial input itself are shown in the first three rows, respectively. Further, the reconstruction of the whitened representation is shown in the fourth row and at the last row; the reconstruction of the refined representation is demonstrated where the noise seems to be neutralized almost completely.

\end{document}